\documentclass[conference]{IEEEtran}
\IEEEoverridecommandlockouts
\usepackage{cite}
\usepackage{amsmath,amssymb,amsfonts}
\usepackage{algorithmic}
\usepackage{graphicx}
\usepackage{textcomp}
\usepackage{xcolor}
\usepackage{booktabs}
\def\BibTeX{{\rm B\kern-.05em{\sc i\kern-.025em b}\kern-.08em
    T\kern-.1667em\lower.7ex\hbox{E}\kern-.125emX}}
    
\usepackage{url}
\usepackage{bbm}

\usepackage{fancyhdr}

\usepackage[hidelinks]{hyperref}
\hypersetup{
  colorlinks   = true, 
  urlcolor     = blue, 
  citecolor   = green 
}
    
\begin{document}


\title{Leveraging the Learnable Vertex-Vertex Relationship to Generalize Human Pose and Mesh Reconstruction for In-the-Wild Scenes}


\author{\IEEEauthorblockN{Trung Quang Tran\IEEEauthorrefmark{1}, Cuong Cao Than\IEEEauthorrefmark{1}, Hai Thanh Nguyen\IEEEauthorrefmark{1}, and
Hong Hoang Si\IEEEauthorrefmark{2}}
\\
\IEEEauthorblockA{\IEEEauthorrefmark{1}Research and Development Team, Asilla Inc., Hanoi, Vietnam \\
\IEEEauthorrefmark{2}School of Electrical and Electronic Engineering (SEEE), Hanoi University of Science and Technology, Hanoi, Vietnam \\
Email: \textit{trungtq@asilla.net}, \textit{cuong@asilla.net}, \textit{hai@asilla.net},  \textit{hong.hoangsy@hust.edu.vn}
}}
 
\maketitle

\thispagestyle{fancy}
\pagestyle{fancy}
\fancyhf{}
\lhead{The 9th NAFOSTED Conference on Information and Computer Science (NICS 2022)}
\rhead{October 31-November 01, 2022}
\cfoot{\thepage}

\begin{abstract}
We present MeshLeTemp, a powerful method for 3D human pose and mesh reconstruction from a single image. In terms of human body priors encoding, we propose using a learnable template human mesh instead of a constant template as utilized by previous state-of-the-art methods. The proposed learnable template reflects not only vertex-vertex interactions but also the human pose and body shape, being able to adapt to diverse images. We conduct extensive experiments to show the generalizability of our method on unseen scenarios.
\end{abstract}


\section{Introduction}

Reconstructing 3D human pose and mesh from a single image can be categorized into two main approaches. The first approach is referred to as the parametric one that aims to predict pose and shape parameters \cite{kanazawa2018end,kocabas2020vibe}. 
These methods then use a parametric model like SMPL \cite{loper2015smpl} to generate the 3D human mesh. Since the parametric models incorporate strong prior knowledge about human shape, the parametric approach is robust with various conditions and very little data. Unfortunately, this approach highly depends on the parametric models which are built out of particular exemplars. The second approach, a non-parametric one, has emerged as a straightforward yet effective way to directly predict 3D coordinates of human pose and mesh \cite{sun2018integral,kolotouros2019convolutional,lin2021end,lin2021mesh}. This paper focuses on elaborating on non-parametric methods because the main part of our method is based on non-parametric ones. Among non-parametric methods, a variety of advanced architectures have been leveraged to model vertex-vertex interactions. While GraphCMR \cite{kolotouros2019convolutional} and METRO \cite{lin2021end} use Graph Convolutional Neural Networks (GCNNs) and Transformers respectively, Mesh Graphormer \cite{lin2021mesh} utilizes both these two architectures. These three methods introduce a template human mesh to preserve the positional information, but only a constant template is used.

\begin{figure}[htbp]
\centerline{\includegraphics[width=0.8\linewidth]{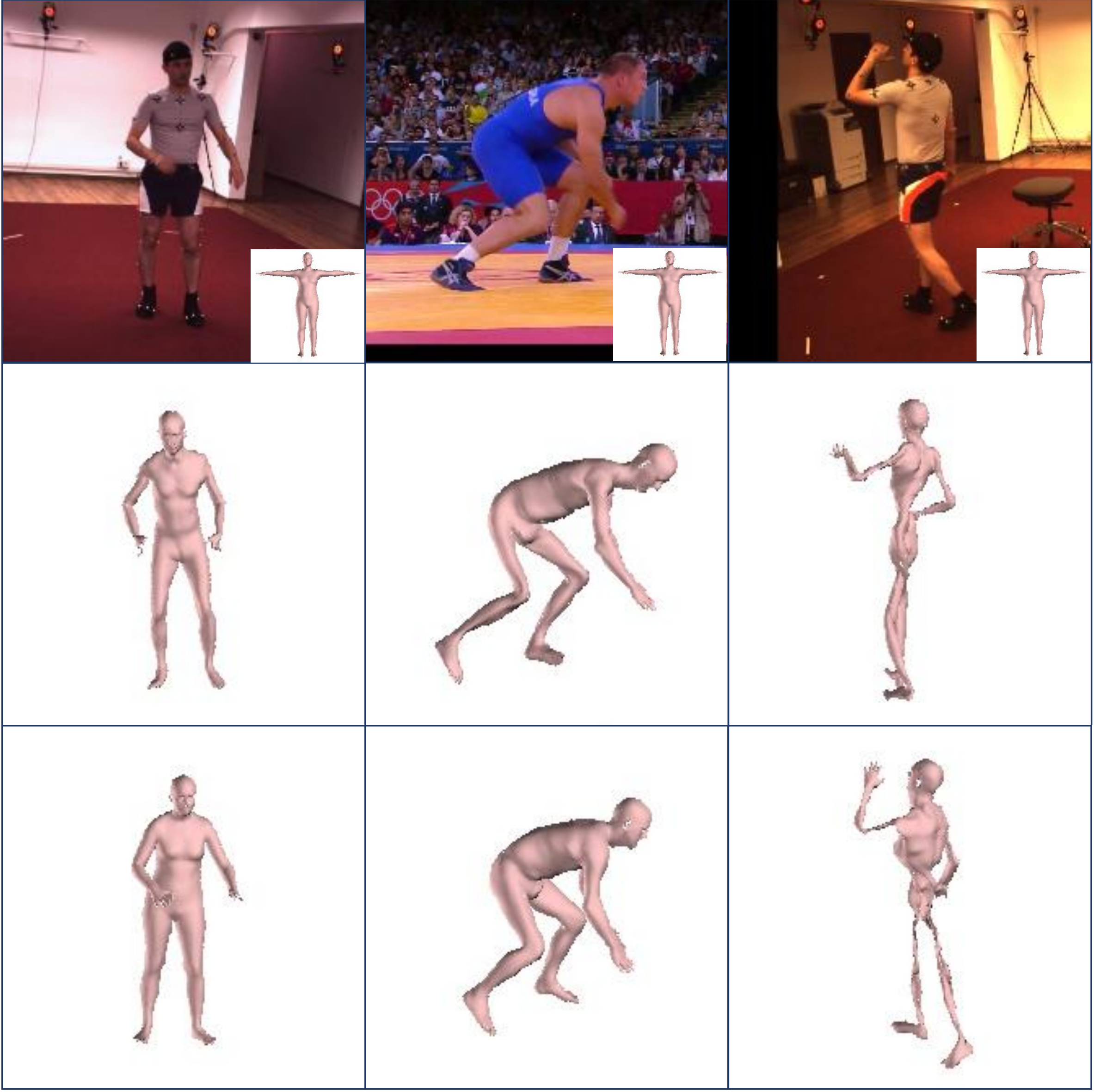}}
\caption{Illustrations of learnable template human mesh (the 2$^{nd}$ and 3$^{rd}$ row) compared to constant template (the small box in the 1$^{st}$ row). While the 2$^{nd}$ row shows the learnable template at the initial training steps, the 3$^{rd}$ row is for the later training steps.}
\label{fig_motivation}
\end{figure}

\begin{figure*}[htbp]
\centerline{\includegraphics[width=0.9\linewidth]{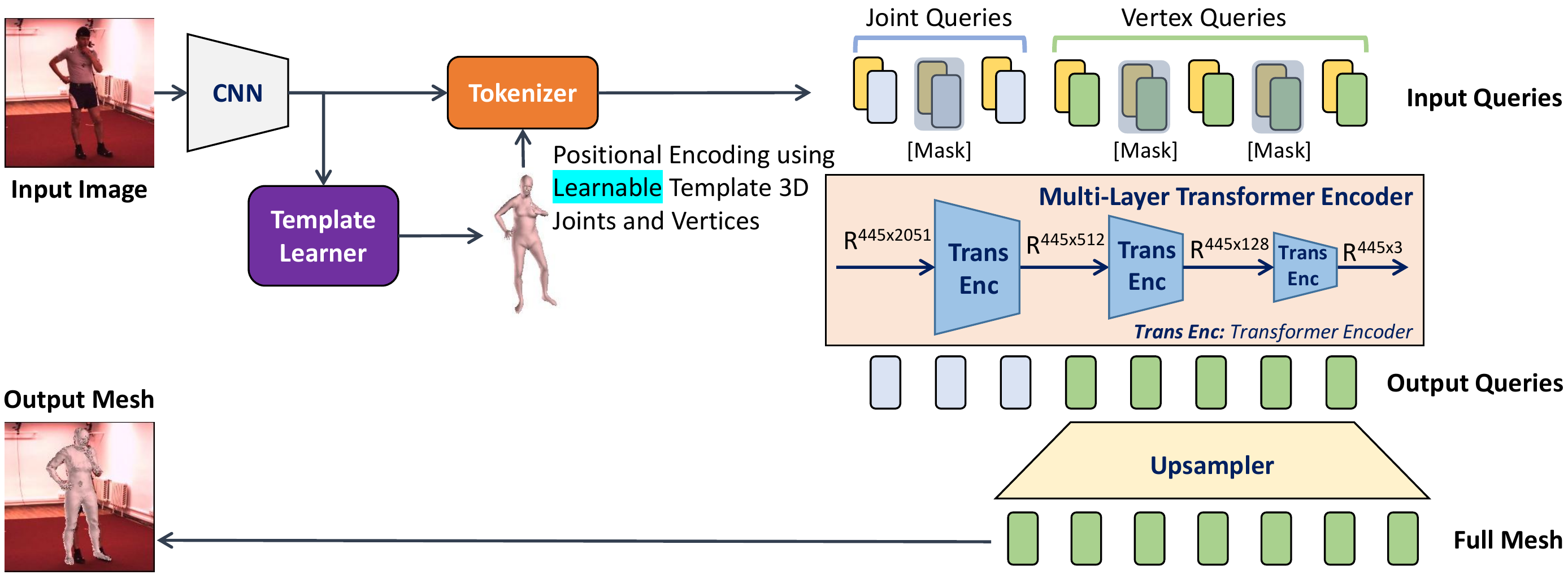}}
\caption{MeshLeTemp for 3D Human Pose and Mesh Reconstruction. Our framework first uses a feature extractor which is a Convolutional Neural Network (CNN) to extract the image features. These image features are fed into a template learner to obtain the template human mesh. The learnable template mesh and the extracted image features are then combined into a set of joint queries and vertex queries. Finally, a Multi-Layer Transformer Encoder takes as input the set of queries to reconstruct the 3D human pose and mesh.}
\label{fig_diagram}
\end{figure*}

In this paper, we propose MeshLeTemp, a multi-layer Transformers-based method, leveraging the \textbf{learnable} vertex-vertex relationship to effectively reconstruct 3D human pose and mesh. The intuition behind injecting a template human mesh into the image features is to embed the positional knowledge in the extracted image features. The constant template might be enough for encoding vertex-vertex connections such as eye-ear or nose-mouth connections. However, it does not reflect the human pose and body shape when feeding different images. We argue that the template human mesh should contain information about not only the vertex-vertex interactions but also the human pose and body shape. We thereby propose leveraging a learnable template human mesh, as illustrated in Fig. \ref{fig_motivation}. While the constant template is the same for all images, our learnable template is able to adapt to diverse human poses. Our template first learns the body pose in the initial training steps and then obtains the knowledge of body shape in the later ones. When the body shape is learned, the body pose is fine-tuned accordingly to achieve a more sophisticated template. For the hard pose occurring when the person turns around (the 3$^{rd}$ column of Fig. \ref{fig_motivation}), the body pose and shape are harder to learn. While the body pose fits the whole body, the body shape prioritizes fitting the upper body. Our experimental results on in-the-wild datasets, 3DPW and MPI-INF-3DHP, proved the efficacy of the learnable vertex-vertex relationship in learning 3D human pose and mesh. To the best of our knowledge, we are the first to utilize the learnable vertex-vertex interactions to support 3D human pose and mesh reconstruction.

Our key contributions are summarized as follows:
\begin{itemize}
    \item We propose a powerful method, MeshLeTemp, taking the advantage of the learnable template human mesh to reconstruct 3D human pose and mesh effectively.
    \item MeshLeTemp achieves a better generalization compared to previous state-of-the-art methods.
\end{itemize}

\section{Related Work}

\textbf{Human Mesh Reconstruction.} \space Human mesh reconstruction has attracted the attention of researchers in recent years. The impressive reconstructions can be obtained by using physical devices such as motion cameras~\cite{ionescu2013human3} or Inertial Measurement Unit (IMU) motion sensors~\cite{von2018recovering}. This approach is costly and even requires complex algorithms to process the sensor's outputs. Therefore, software-based methods have emerged as a promising approach in this field. The software-based approach can be divided into two main branches. The first branch utilizes parametric models such as SMPL~\cite{loper2015smpl} 
or STAR~\cite{osman2020star} to generate the 3D mesh from predicted parameters. This branch takes the advantage of the prior knowledge about human shape incorporated into the parametric models. 
The parametric methods strongly depend on the parametric models and are limited by particular exemplars. Therefore, the second branch of the software-based approach, including non-parametric methods, aims to directly predict the 3D human pose and mesh from a monocular image. 

Recently, Graph Convolutional Neural Networks (GCNNs) and Transformers 
have proven their efficacy in 3D human pose and mesh reconstruction \cite{kolotouros2019convolutional,lin2021end,lin2021mesh}. GraphCMR \cite{kolotouros2019convolutional} utilized GCNNs to model local vertex-vertex interactions but did not consider non-local interactions which also have strong correlations. To overcome this limitation, METRO \cite{lin2021end} proposed a simple yet effective framework, that utilizes Transformers, to model both local and non-local vertex-vertex interactions. Mesh Graphormer \cite{lin2021mesh} injected graph convolutions into Transformers to further improve local interactions. Our method leverages Multi-Layer Transformer Encoder for 3D human pose and mesh reconstruction.

\textbf{Human Body Priors Encoding.} \space Encoding human body priors to make 3D human pose and mesh reconstruction more robust is one of the important techniques which has been attractive recently. 
GraphCMR \cite{kolotouros2019convolutional} proposed to attach the image feature vector to a 3D template human mesh. This embedded template human mesh was then passed to a series of graph convolutional layers to regress the 3D vertex coordinates. Similarly, METRO \cite{lin2021end} obtained positional encoding by concatenating the template human mesh and the image features. However, instead of using Graph Convolutional Neural Networks, METRO utilized Transformers to model vertex-vertex and vertex-joint interactions. Graphormer \cite{lin2021mesh} is the most recent method which injected graph convolutions into the transformer blocks to make local and global interaction modeling more robust. Additionally, Graphormer extracted the grid features from the last convolutional block of the feature extractor and used it to obtain fine-grained local details.

\section{MeshLeTemp}
\label{sec_method}

This section is organized as follows. Section \ref{sec_transformer} presents the overall framework of MeshLeTemp especially Multi-Layer Transformer Encoder, one of the important parts of our method to predict the 3D human joints and mesh simultaneously. Next, Section \ref{sec_temp} describes our proposed block to learn the template human mesh. 
Finally, Section \ref{sec_training_details} presents our training details.

\subsection{Overall Framework}
\label{sec_transformer}

The overall framework of MeshLeTemp is illustrated in Fig. \ref{fig_diagram}. Our model consists of two main parts. While the first part uses a Convolutional Neural Network (CNN) to extract the image feature vector, the second one is responsible for generating 3D human joints and mesh by utilizing a Multi-Layer Transformer Encoder (MTE). We leverage HRNets architecture~\cite{wang2020deep}, an existing large-scale network, for extracting the image features. HRNets have been proven to be a powerful architecture in visual recognition. The output of CNN is 2048 feature maps with the size of $7\times 7$ each. These feature maps are then used by a tokenizer and a template learner to generate 445 input queries, including 14 joint queries and 431 vertex ones. Each query either joint or vertex is a 2051-dimensional vector, where 2048 elements are for image features and 3 remaining ones are 3D coordinates. More details about the tokenizer and template learner will be presented in Section \ref{sec_temp}.

As illustrated in Fig. \ref{fig_diagram}, Multi-Layer Transformer Encoder (MTE) takes as input joint queries and vertex queries. For joint queries, we train our model with 14 keypoints which in order are right ankle, right knee, right hip, left hip, left knee, left ankle, right wrist, right elbow, right shoulder, left shoulder, left elbow, left wrist, neck, and head. For vertex queries, as recommended by Lin \textit{et al.}~\cite{lin2021end,lin2021mesh}, we use a coarse human mesh, containing 431 vertices, to make the training faster. MTE consists of three transformer encoder blocks with dimensionality gradually reduced. The output of MTE includes 3D coordinates for 431 vertices and 14 keypoints. At the end of the framework, MeshLeTemp upsamples the predicted coarse mesh to the original human mesh (6890 vertices) by using learnable Multi-Layer Perceptions (MLPs) layers. To take the occlusion into account, we utilize Masked Vertex Modeling (MVM) which was successfully used by METRO~\cite{lin2021end}. MVM masks some percentages of the input queries randomly. Instead of recovering the masked inputs as done by Masked Language Modeling (MLM) \cite{devlin2018bert}, MTE is asked to regress all the joints and vertices.

\subsection{Learnable Template Human Mesh}
\label{sec_temp}

\begin{figure}[htbp]
\centerline{\includegraphics[width=0.9\linewidth]{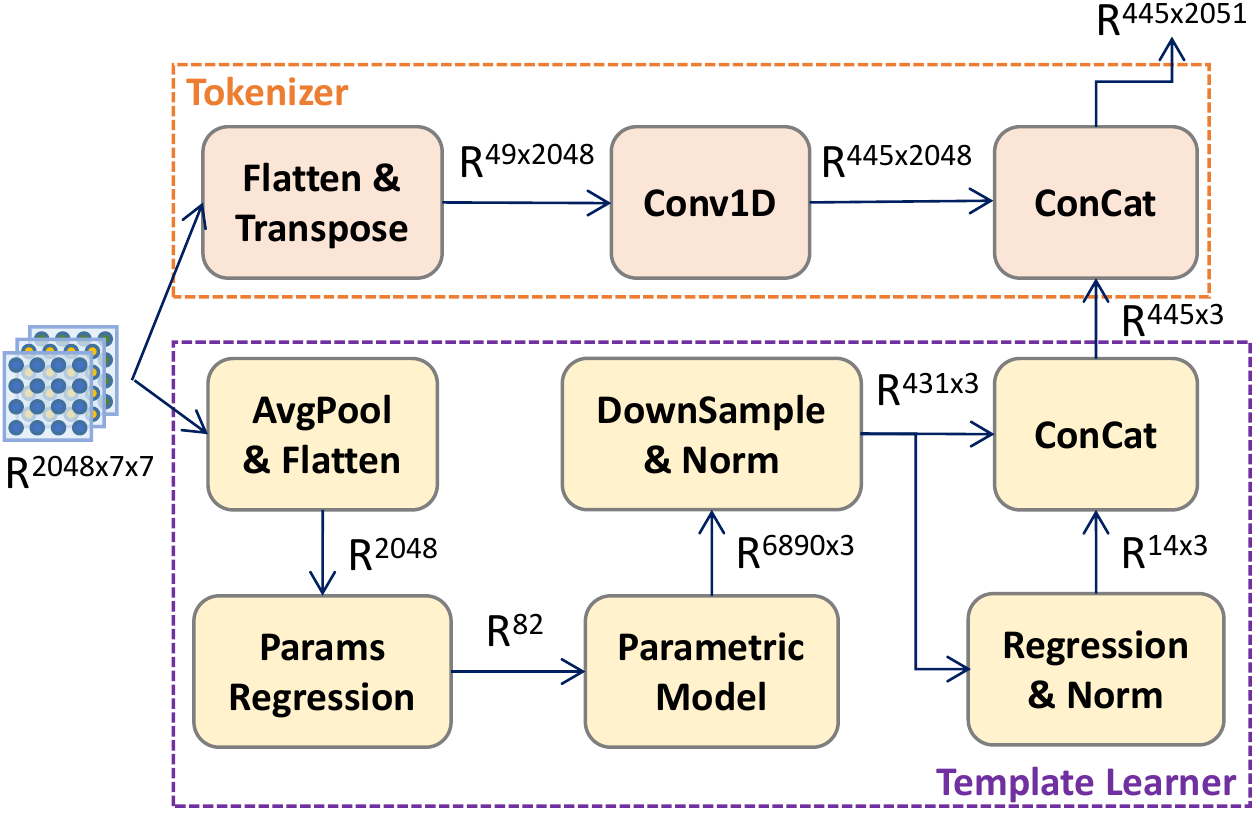}}
\caption{Tokenizer and Template Learner for preparing joint and vertex queries. Especially, Template Learner is responsible for learning template human mesh from extracted image features.}
\label{fig_learnable_temp}
\end{figure}

In the same spirit as positional encoding \cite{kolotouros2019convolutional,lin2021end,lin2021mesh}, we utilize a template human mesh to preserve the positional information of vertex-vertex interactions. Previous works used a constant template which is the same for all input images. As a result, this positional encoding does not reflect the human pose and body shape. For example, the standing pose and bowing pose will use the same template human mesh, as described in Fig. \ref{fig_motivation}. We argue that it would be beneficial if we also consider the human pose and body shape for positional encoding. We thereby propose an extra block, called Template Learner (TL), to learn the template human mesh from the corresponding image features, as illustrated in Fig. \ref{fig_learnable_temp}. The extracted image features, $F\in\mathbb{R}^{2048\times 7\times 7}$, go through an Average Pooling layer and are then flattened to a 2048-dimensional vector. At the heart of Template Learner, we use Multi-Layer Perceptions (MLPs) layers, namely Parameters Regression, to regress the body and shape parameters, $\theta\in\mathbb{R}^{82}$. These parameters are then fed into a parametric model, which is SMPL~\cite{loper2015smpl} in our work, to obtain the learnable template human mesh, $V^{temp}_{3D}\in\mathbb{R}^{6890\times 3}$. We do not directly use $V^{temp}_{3D}$ but downsample it to $\overline{V}^{temp}_{3D}\in\mathbb{R}^{431\times 3}$. Notably, the 3D joints can be obtained from the predicted 3D vertices by using a pre-defined regression matrix $G\in\mathbb{R}^{K\times M}$, as shown in the literature~\cite{choi2020pose2mesh,kanazawa2018end,kolotouros2019convolutional}, where $K$ is the number of joints and $M$ is the number of vertices of a person. We are thereby able to obtain the regressed 3D joints, $J^{temp}_{3D}\in\mathbb{R}^{14\times 3}$, from $\overline{V}^{temp}_{3D}$. The proposed learnable template is robust with various input images, which will be clarified in Section \ref{sec_experimental_results}.

As shown in Fig. \ref{fig_learnable_temp}, $\overline{V}^{temp}_{3D}$ and $J^{temp}_{3D}$ will be used by Tokenizer to create input queries for Multi-Layer Transformer Encoder. Tokenizer takes as input the image features $F$ to form a set of base queries, $\overline{Q}=\{\overline{q}^{J}_{i}\in\mathbb{R}^{2048}:i\in(1,...,14)\}\cup\{\overline{q}^{V}_{j}\in\mathbb{R}^{2048}:j\in(1,...,431)\}$. $\overline{q}^{J}_i$ is concatenated with the corresponding 3D coordinates of $J^{temp}_{3D}$ to obtain the complete joint query $q^{J}_{i}\in\mathbb{R}^{2051}$. Similarly, we combine $\overline{q}^{V}_{j}$ and $\overline{V}^{temp}_{3D}$ to obtain the complete vertex query $q^{V}_{j}\in\mathbb{R}^{2051}$. Finally, we have 445 input queries, where each query is either joint query $q^{J}_{i}$ or vertex query $q^{V}_{j}$, as mentioned in Section \ref{sec_transformer}.

\subsection{Training Details}
\label{sec_training_details}

\begin{table*}[htbp]
\caption{Generalizability of the methods on 3DPW and MPI-INF-3DHP dataset. The results of METRO and Graphormer are produced using the provided checkpoints.}
\begin{center}
\begin{tabular}{lcccccccc}
\toprule
     & \multicolumn{3}{c}{3DPW val} & \multicolumn{3}{c}{3DPW test} & \multicolumn{2}{c}{MPI-INF-3DHP} \\
\cmidrule(r){1-1}\cmidrule(r){2-4}\cmidrule(r){5-7}\cmidrule{8-9}
Method & MPVE $\downarrow$ & MPJPE $\downarrow$ & PA-MPJPE $\downarrow$ & MPVE $\downarrow$ & MPJPE $\downarrow$ & PA-MPJPE $\downarrow$ & MPJPE $\downarrow$ & PA-MPJPE $\downarrow$ \\
\cmidrule(r){1-1}\cmidrule(r){2-4}\cmidrule(r){5-7}\cmidrule{8-9}
METRO~\cite{lin2021end} & 127.5 & 112.4 & 71.0 & 130.2 & 114.3 & 67.2 & 118.2 & 71.7 \\
Graphormer~\cite{lin2021mesh} & 147.6 & 131.1 & 86.7 & 157.6 & 142.2 & 88.0 & 121.4 & 72.5 \\
\cmidrule(r){1-1}\cmidrule(r){2-4}\cmidrule(r){5-7}\cmidrule{8-9}
MeshLeTemp (\textbf{Ours}) & \textbf{118.3} & \textbf{102.0} & \textbf{66.0} & \textbf{123.4} & \textbf{106.0} & \textbf{62.8} & \textbf{113.2} & \textbf{68.9} \\
\bottomrule
\end{tabular}
\label{table_res_generalization_3dpw}
\end{center}
\end{table*}

We use a similar training strategy as used in METRO~\cite{lin2021end} and Graphormer~\cite{lin2021mesh}. Concretely, we use $L_1$ loss for both predicted 3D vertices and predicted 3D joints, leading to $\mathcal{L}_V$ and $\mathcal{L}_J$ respectively. The 3D joints can also be obtained from the predicted 3D vertices by using a pre-defined regression matrix, as presented in Section \ref{sec_temp}. Therefore, we use $L_1$ loss for these regressed 3D joints, resulting in $\mathcal{L}_J^{reg}$. On top of the outputs of the model, we use Multi-Layer Perceptions (MLPs) layers to learn the camera parameters, which are used to project the 3D joints to the 2D space. $L_1$ loss is used for these projected 2D joints, leading to $\mathcal{L}_J^{proj}$. Different from previous methods using the constant template human mesh, we additionally utilize $L_1$ loss to supervise the learnable template which is built out of 3D vertices, resulting in $\mathcal{L}_V^{temp}$. Our overall objective function is formulated as follows:
\begin{equation}
    \mathcal{L}=\alpha\times (\mathcal{L}_V+\mathcal{L}_J+\mathcal{L}_J^{reg}) + \alpha_{temp}\times \mathcal{L}_V^{temp} + \beta\times\mathcal{L}_J^{proj},
\end{equation}
where $\alpha$, $\alpha_{temp}$, and $\beta$ are scalar hyperparameters denoting the weights of the loss elements.

We train MeshLeTemp with the Adam optimizer \cite{kingma2014adam} and cosine learning rate scheduler \cite{loshchilov2016sgdr}. Given a base learning rate $\eta$, the learning rate at the training step $s$ is set to $\eta\times\cos{\frac{7\pi s}{16S}}$, where $S$ is the total number of training steps. In our experiments, we use a base learning rate of $1e-4$ and train our model for $50$ epochs. Notably, our training is much faster than the two most relevant methods, METRO and Graphormer, which trained their models for $200$ epochs.

\section{Experimental Results}
\label{sec_experimental_results}

We first show that MeshLeTemp obtains a better generalization compared to other methods. The comparison with state-of-the-art methods is also provided. Finally, we conduct ablation studies, including qualitative results, to show the impact of the learnable template human mesh.

\subsection{Datasets}
\label{sec_exp_dataset}

We conduct extensive experiments with the mix-training strategy using 3D and 2D data. We use public datasets including Human3.6M~\cite{ionescu2013human3}, UP-3D~\cite{lassner2017unite}, MuCo-3DHP~\cite{mehta2018single}, COCO~\cite{lin2014microsoft}, and MPII~\cite{andriluka20142d}. For Human3.6M dataset, we use the pseudo-labels generated by SMPLify-X~\cite{pavlakos2019expressive}. In the common setting, we use the subjects S1, S5, S6, S7, and S8 for training, and keep the subjects S9 and S11 for testing. For a fair comparison with other methods, we also use 3DPW~\cite{von2018recovering} dataset for training and testing. We additionally conduct evaluation on MPI-INF-3DHP \cite{mono-3dhp2017} dataset and visualization on SSP-3D \cite{sengupta2020synthetic} dataset. 

\subsection{Generalization to In-the-Wild Datasets}
\label{sec_exp_generalization}


We check the generalizability of our model by comparing it with the two most relevant methods, METRO~\cite{lin2021end} and Graphormer~\cite{lin2021mesh}, as shown in Table \ref{table_res_generalization_3dpw}. For a fair comparison, we train our model on the same datasets used by METRO and Graphormer. Specifically, we train our model using the mix-training strategy which uses both 3D and 2D training data. The training dataset includes Human3.6M, UP-3D, MuCo-3DHP, COCO, and MPII. As shown in Table \ref{table_res_generalization_3dpw}, MeshLeTemp outperforms METRO and Graphormer on all datasets. For instance, MeshLeTemp improves MPVE by 9.2 and 29.3 points on 3DPW validation set compared to METRO and Graphormer, respectively. On 3DPW test set, MeshLeTemp obtains PA-MPJPE improvement of 4.4 and 25.2 points compared to METRO and Graphormer, respectively. The improvement is also expressed on MPI-INF-3DHP dataset. Notably, both 3DPW and MPI-INF-3DHP are in-the-wild datasets containing complex outdoor scenes, and they were not seen during the training of all methods. As a result, we can argue that MeshLeTemp achieves a much better generalization compared to METRO and Graphormer, especially in in-the-wild scenarios.

\begin{table}[htbp]
\caption{Catastrophic forgetting of the methods on Human3.6M dataset. The results of METRO and Graphormer are produced using the provided checkpoints.}
\begin{center}
\begin{tabular}{lcc}
\toprule
Method & MPJPE $\downarrow$ & PA-MPJPE $\downarrow$ \\
\cmidrule(r){1-1}\cmidrule(r){2-3}
METRO~\cite{lin2021end} & 78.6 (\textit{24.6}) & 48.9 (\textit{12.2}) \\
Graphormer~\cite{lin2021mesh} & 76.1 (\textit{24.9}) & 47.6 (\textit{13.1}) \\
\cmidrule(r){1-1}\cmidrule(r){2-3}
MeshLeTemp (\textbf{Ours}) & \textbf{75.6} (\textit{\textbf{18.8}}) & \textbf{47.4} (\textit{\textbf{9.8}}) \\
\bottomrule
\end{tabular}
\label{table_res_generalization_h36m}
\end{center}
\end{table}

In neural networks, catastrophic forgetting is a critical phenomenon, where the model could obtain high performance on a new task, but its performance might be disastrously degraded on the old task. MeshLeTemp showed the good generalization on 3DPW and MPI-INF-3DHP dataset. We additionally examine the catastrophic forgetting of our method on the old task, which is Human3.6M. To do so, we fine-tune our model on 3DPW training set and evaluate it on Human3.6M, as shown in Table \ref{table_res_generalization_h36m}. Both METRO, Graphormer, and MeshLeTemp are trained and fine-tuned using the same datasets. The number in the parentheses is the performance degradation on Human3.6M when the model is fine-tuned on 3DPW dataset. MeshLeTemp still achieves the better results compared to METRO and Graphormer. Especially, our method has the minimum performance degradation compared to METRO and Graphormer.

\subsection{Comparison with State-of-the-Art Methods}
\label{sec_exp_sota_comparison}

\begin{table*}[htbp]
\caption{Comparison with state-of-the-art methods on Human3.6M and 3DPW dataset. * denotes the results cited from METRO.}
\begin{center}
\begin{tabular}{lccccc}
\toprule
    & \multicolumn{2}{c}{Human3.6M} & \multicolumn{3}{c}{3DPW test} \\
\cmidrule(r){1-1}\cmidrule(r){2-3}\cmidrule(r){4-6}
Method & MPJPE $\downarrow$ & PA-MPJPE $\downarrow$ & MPVE $\downarrow$ & MPJPE $\downarrow$ & PA-MPJPE $\downarrow$ \\
\cmidrule(r){1-1}\cmidrule(r){2-3}\cmidrule(r){4-6}
HMR~\cite{kanazawa2018end} & - & 56.8 & - & - & 81.3* \\
GraphCMR~\cite{kolotouros2019convolutional} & - & 50.1 & - & - & 70.2* \\
SPIN~\cite{kolotouros2019learning} & - & 41.1 & - & - & 59.2 \\
Pose2Mesh~\cite{choi2020pose2mesh} & 64.9 & 46.3 & 105.3 & 89.5 & 56.3 \\
I2LMeshNet~\cite{moon2020i2l} & 55.7 & 41.7 & - & 93.2 & 57.7 \\
VIBE~\cite{kocabas2020vibe} & 65.9 & 41.5 & 99.1 & 83.0 & 52.0 \\
STRAPS~\cite{sengupta2020synthetic} & - & 55.4 & - & - & 66.8 \\
METRO~\cite{lin2021end} & \underline{54.0} & \underline{36.7} & 88.2 & 77.1 & 47.9 \\
Graphormer~\cite{lin2021mesh} & \textbf{51.2} & \textbf{34.5} & \underline{87.7} & \textbf{74.7} & \textbf{45.6} \\
EFT~\cite{joo2021exemplar} & - & 44.0 & - & - & 51.6 \\
HybrIK~\cite{li2021hybrik} & 54.4 & \textbf{34.5} & 94.5 & 80.0 & 48.8 \\
\cmidrule(r){1-1}\cmidrule(r){2-3}\cmidrule(r){4-6}
MeshLeTemp (\textbf{Ours}) & 54.9 & 37.0 & \textbf{86.5} & \underline{74.8} & \underline{46.8} \\
\bottomrule
\end{tabular}
\label{table_sota_comparison}
\end{center}
\end{table*}

We also compare MeshLeTemp with the state-of-the-art methods, as shown in Table \ref{table_sota_comparison}. Notably, we fine-tune our model on 3DPW training set when conducting the comparison on 3DPW dataset, as similar to other methods. The results show that MeshLeTemp strongly outperforms many previous state-of-the-art methods. For instance, MeshLeTemp reduces PA-MPJPE by 4.5 and 5.2 points compared to VIBE on Human3.6M and 3DPW test set, respectively. In comparison with METRO and Graphormer, we obtain the comparable results. On 3DPW dataset, MeshLeTemp has the best result for MPVE and the second-best result for MPJPE and PA-MPJPE. However, the results of MeshLeTemp on Human3.6M are not good compared to METRO and Graphormer. We argue that Human3.6M contains indoor scenarios and simple poses, so the models might be prone to fit this dataset. By using a learnable template instead of a constant one, MeshLeTemp is able to avoid overfitting while obtaining the generalization in outdoor scenarios, as elaborated in Section \ref{sec_exp_generalization}.

\subsection{Impact of the Learnable Template Human Mesh}
\label{sec_exp_template}

As mentioned in Section \ref{sec_training_details}, we use $L_1$ loss to supervise the learnable template human mesh. Table \ref{table_res_impact_template} shows the performance change when varying the loss weight of the learnable template. The loss weight, $\alpha_{temp}$, is a hyperparameter to indicate the contribution of the learnable template to the overall objective function. We start to set $\alpha_{temp}$ to 0.1 and gradually increase it by the step of 0.1. The performance is improved with respect to the larger value of $\alpha_{temp}$. However, the performance is slightly degraded when $\alpha_{temp}$ reaches 0.4. Therefore, we try a value between 0.3 and 0.4. We found that $\alpha_{temp}$ of 0.33 gives the best result, though the gap is not significant. We can conclude that the performance can be improved if the learnable template contributes much to the overall objective function.

\begin{table}[htbp]
\caption{Ablation study of the learnable template human mesh, evaluated on 3DPW dataset.}
\begin{center}
\begin{tabular}{lrrrrr}
\toprule
$\alpha_{temp}$ & 0.10 & 0.20 & 0.30 & 0.33 & 0.40 \\
\cmidrule(r){1-1}\cmidrule(r){2-6}
PA-MPJPE $\downarrow$ & 71.7 & 65.1 & 64.4 & \textbf{64.0} & 64.9 \\
\bottomrule
\end{tabular}
\label{table_res_impact_template}
\end{center}
\end{table}

\begin{figure}[htbp]
\centerline{\includegraphics[width=0.95\linewidth]{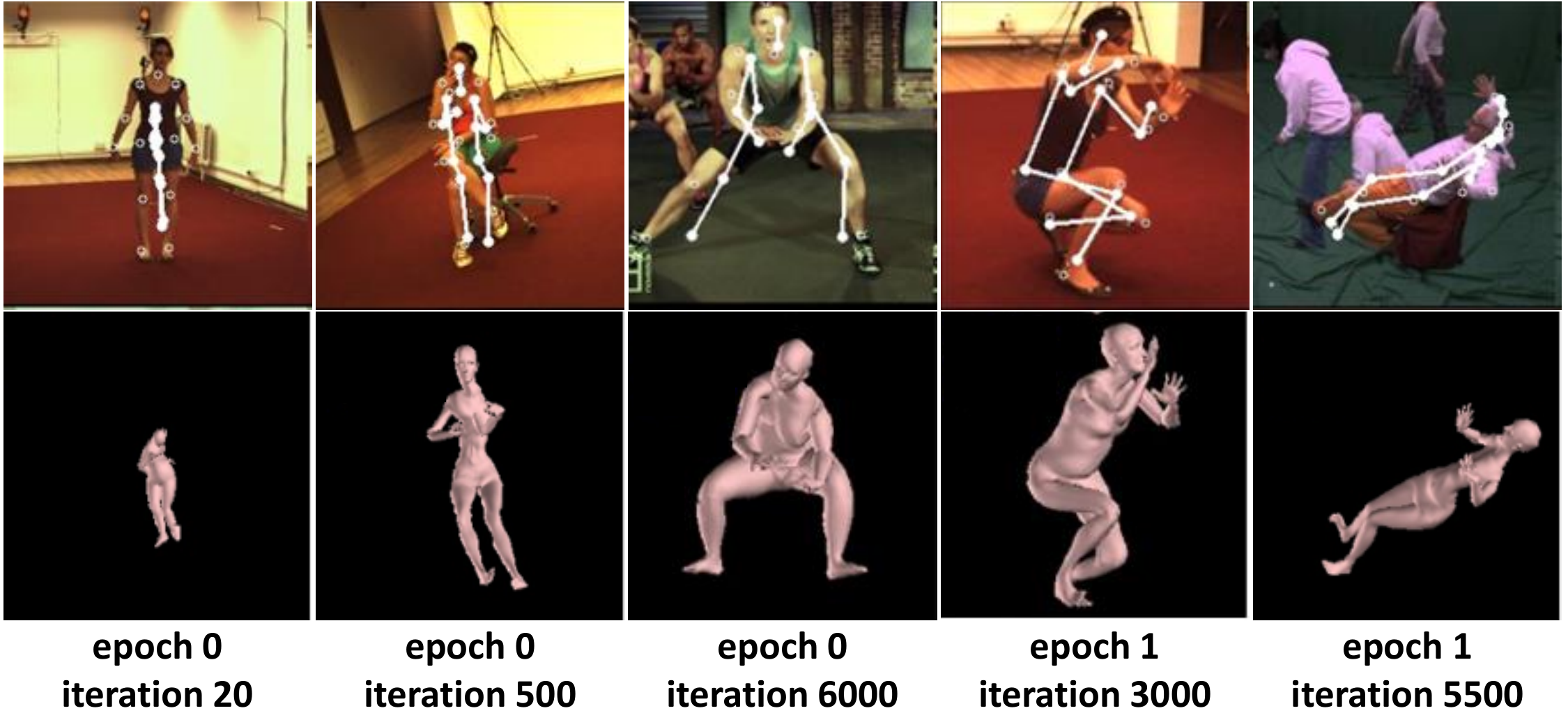}}
\caption{Illustrations of the learnable template human mesh during the training process.}
\label{fig_template_progress}
\end{figure}

Additionally, we observe the learnable template during the training process to see how the template human mesh progresses, as illustrated in Fig. \ref{fig_template_progress}. The top row depicts the predicted 2D keypoints, and the bottom row is for the learned 3D template human mesh.
\begin{itemize}
    \item At the first iterations (the 1$^{st}$ column for instance), the templates are quite similar among images.
    \item When the training progresses, the template starts to fit the specific images but is still poor at hard poses (the 2$^{nd}$ and 3$^{rd}$ column).
    \item Finally, the visualization in the 4$^{th}$ and 5$^{th}$ column shows that the template human mesh tries to handle hard poses.
\end{itemize}

\subsection{Qualitative Results}
\label{sec_exp_qualitative_result}

\begin{figure}[htbp]
\centerline{\includegraphics[width=1.0\linewidth]{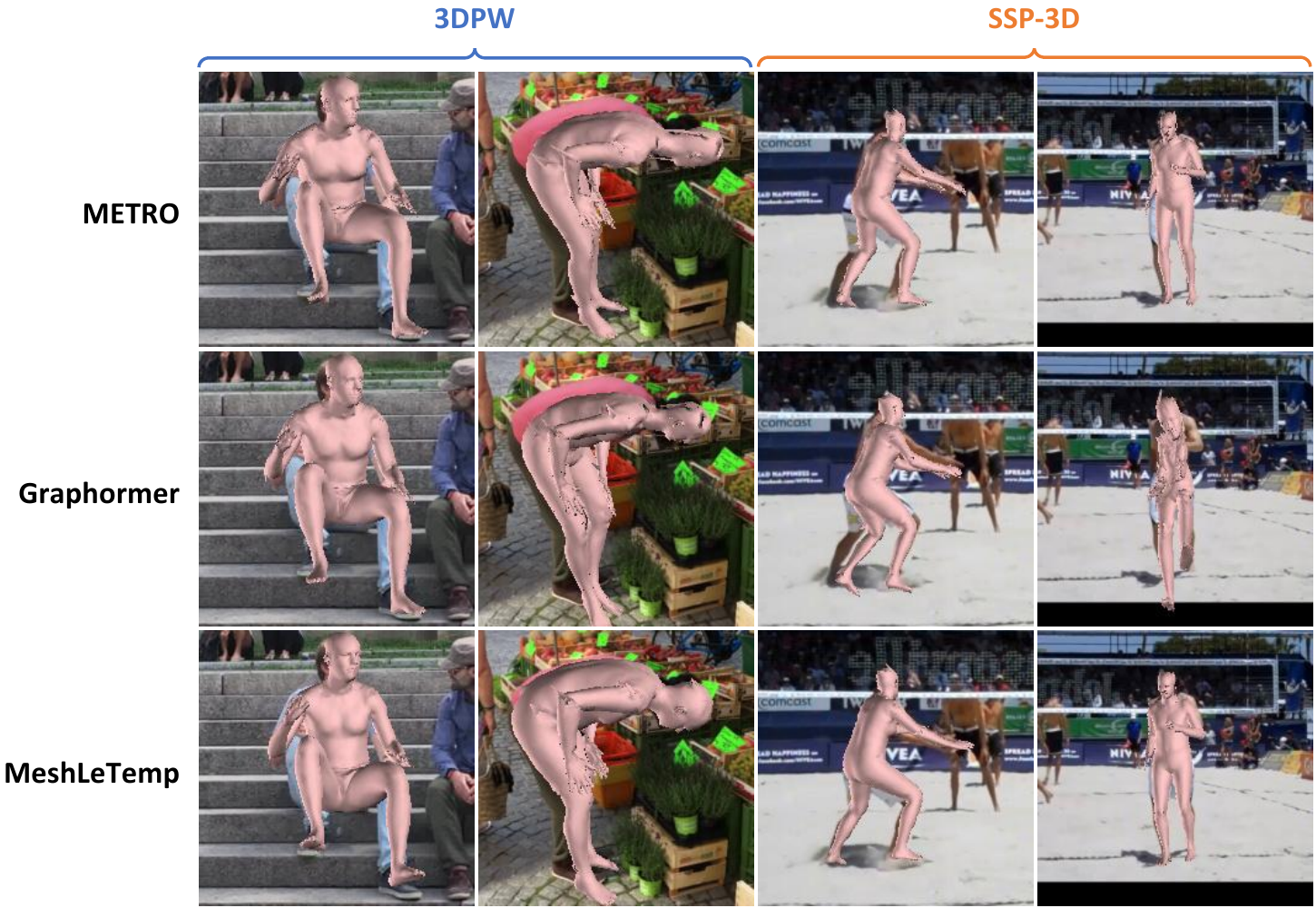}}
\caption{Qualitative results of the methods on in-the-wild datasets. All models are not trained on 3DPW and SSP-3D dataset.}
\label{fig_vis}
\end{figure}

Fig. \ref{fig_vis} shows the qualitative results of our MeshLeTemp compared to the two most relevant methods, METRO and Graphormer. We evaluate the models on in-the-wild images, 3DPW \cite{von2018recovering} and SSP-3D \cite{sengupta2020synthetic}, to check the generalizability of the methods.
\begin{itemize}
    \item For the simple case where the background is not complicated, all methods give acceptable results, as illustrated in the 1$^{st}$ column of Fig. \ref{fig_vis}. 
    \item When the background becomes harder as shown in the 2$^{nd}$ column of Fig. \ref{fig_vis}, MeshLeTemp outperforms METRO and Graphormer. Both METRO and Graphormer fail to fit the reconstructed 3D mesh to the person (the backbone deviates, and the body shape is not matched). Otherwise, MeshLeTemp can fit the reconstructed 3D mesh to the person well.
    \item On SSP-3D dataset where people play sport, MeshLeTemp performs better than METRO and Graphormer. Especially, Graphormer seems to be unstable when it fails to reconstruct the 3D mesh for the 4$^{th}$ column of Fig. \ref{fig_vis}.
\end{itemize}

\section{Conclusion}

We introduce a powerful method to effectively encode the human body priors into the image features. Instead of using the constant template human mesh as previous methods did, our method, MeshLeTemp, leverages the learnable template to reconstruct 3D human pose and mesh from a single input image. The extensive experiments show that our method obtains the better generalization compared to previous state-of-the-art methods. For future work, we are interested in elaborately adapting our method to another domain such as 3D hand reconstruction.

\bibliographystyle{unsrt}
\bibliography{bibliography}


\end{document}